\documentclass[lettersize,journal]{IEEEtran}
\usepackage{amsmath,amsfonts}
\usepackage{algorithmic}
\usepackage{algorithm}
\usepackage{array}
\usepackage[caption=false,font=normalsize,labelfont=sf,textfont=sf]{subfig}
\usepackage{textcomp}
\usepackage{stfloats}
\usepackage{url}
\usepackage{verbatim}
\usepackage{graphicx}
\usepackage{cite}
\usepackage{xcolor}
\usepackage{booktabs}
\usepackage{multirow}
\usepackage[numbers,sort&compress]{natbib}

\begin{document}

\title{Multi-Agent Based Transfer Learning for Data-Driven Air Traffic Applications}

\author{Chuhao Deng, Hong-Cheol Choi, Hyunsang Park, Inseok Hwang,~\IEEEmembership{Member,~IEEE,}
\thanks{This paper was produced by the IEEE Publication Technology Group. They are in Piscataway, NJ.}
\thanks{Manuscript received April 19, 2021; revised August 16, 2021.}}

\markboth{Journal of \LaTeX\ Class Files,~Vol.~14, No.~8, August~2021}%
{Shell \MakeLowercase{\textit{et al.}}: A Sample Article Using IEEEtran.cls for IEEE Journals}


\maketitle

\begin{abstract}
Research in developing data-driven models for Air Traffic Management (ATM) has gained a tremendous interest in recent years. However, data-driven models are known to have long training time and require large datasets to achieve good performance. To address the two issues, this paper proposes a Multi-Agent Bidirectional Encoder Representations from Transformers (MA-BERT) model that fully considers the multi-agent characteristic of the ATM system and learns air traffic controllers' decisions, and a pre-training and fine-tuning transfer learning framework. By pre-training the MA-BERT on a large dataset from a major airport and then fine-tuning it to other airports and specific air traffic applications, a large amount of the total training time can be saved. In addition, for newly adopted procedures and constructed airports where no historical data is available, this paper shows that the pre-trained MA-BERT can achieve high performance by updating regularly with little data. The proposed transfer learning framework and MA-BERT are tested with the automatic dependent surveillance-broadcast data recorded in 3 airports in South Korea in 2019. 
\end{abstract}

 
\section{Introduction} \label{sec:intro}
\IEEEPARstart{T}{he} Air Traffic Management (ATM) system is not only a highly complex system that contains operations across multiple domains, but also a continuously growing system. In 2021, the international civil aviation organization reports the total number of passengers carried on scheduled services has reached 2.3 billion, a 28.1 percent increase from 2020 \cite{icao}. Such rapid growth induces the construction of new airports and the renovation of existing airports. For instance, federal aviation administration’s airport improvement program that aims to fund airport infrastructure projects such as runways, texiways, airport signage, etc., and the program provides more than 3.18 billion dollars of funds annually to more than 3,300 eligible airports. 

In response to the rapid growth and large-scale renovation, new airports are built and operations in existing airports are often changed. In addition, to maintain the efficiency and safety for the future of the ATM system, extensive research has been conducted, especially in data-driven models due to their ability to model complex systems without prior assumptions on aircraft motions or operations \cite{HCchoi,scad,hc_eta}. However, data-driven models are known to have long training time and require large datasets to achieve desirable performance. Furthermore, when the data distribution is changed, that is, the data-driven models are applied to a different sector or airport, most data-driven models need to be rebuilt from scratch using the newly collected data \cite{tl_survey_1}, which means very long total training time. To address the issues, transfer learning is introduced. In general, transfer learning aims to use the knowledge learned from one source domain to other target domains \cite{tl_survey_1,tl_survey_2}. By doing so, the learning for target domains is quick and requires much less data. A branch of transfer learning is called domain adaptation, which aims to transfer knowledge from a labeled source domain to an unlabeled target domain \cite{tl}. Over the years, various measurements are proposed to be applied to tackle the problem, including maximum mean discrepancy \cite{mmd_1,mmd_2}, Kullback-Leibler divergence \cite{kl_1,kl_2}, and Jensen-Shannon divergence \cite{js_1,js_2}. The underlaying assumption of domain adaptation is that the labels in source and target domains are the same. For instance, transferring a model that classifies images from the Amazon website to classify images collected by webcams. Although the images from two domains have different styles, categories of those images are the same.

In recents years, the growth of computing power and the size of open dataset prompt the development of Large Language Models (LLMs) such as the famous ChatGPT \cite{gpt}. LLMs rely on transfer learning. However, instead of reducing the shifts in data distribution, they rely on pre-training on large datasets to learn the pre-trained representations of data. In general, there are two established approaches for applying pre-trained representations to downstream tasks: feature-based and fine-tuning. The feature-based approach, as pointed out in \cite{bert}, includes pre-trained representations as additional features to task-specific architectures. However, the task-specific architectures make the feature-based approach not ideal for transferring data-driven models to both different airports and tasks. The fine-tuning approach, on the other hand, trains for downstream tasks via fine-tuning the parameters of a pre-trained model. Bidirectional Encoder Representations from Transformers (BERT) is one of the first LLMs and it is based on the fine-tuning approach \cite{bert}. During pre-training, BERT learns the deep representation of natural language. Fine-tuning then maps the deep representation of texts into the dimension of downstream natural language tasks with little data and training time. It is found that pre-training and fine-tuning allow one model to be transferred to many tasks and increases the performance of the model \cite{bert,gpt,tay2021scale}. Since a paragraph can be seen as a sequence and each word contains multiple characters, natural language data can be seen as multi-variate time series data. Therefore, in \cite{multi_trans}, a Transformer is pre-trained on a variety of time series datasets and fine-tuned to both classification and regression tasks. It achieves the state-of-the-art performance not only when the entire dataset is available, but also when just a fraction of the dataset is available, indicating the pre-trained model's ability to quickly adapt to new tasks. 

In addition to the popularity of transfer learning using pre-trained models in natural language, many works also focus on transfer learning in intelligent transportation system. For instance, predicting traffic flow in cities \cite{flow_1,flow_2,flow_3,flow_4}, map-matching \cite{match}, and drivers' behavior in lane-changes \cite{drive_1,drive_2}. For the ATM system, the authors in \cite{av_1} proposed a Long Short-Term Memory (LSTM) based transfer learning model for predicting delays of low cost airlines. The model first combines the data from a big airline with the data from a small airline, and then trains with a novel loss function that contains a weighted loss term for the clusters of the small airline's data. The result shows improvements compared with the baseline LSTM model and the linear regression model. However, the transfer learning scheme in \cite{av_1} is task-specific and it does not help reducing the total training time. 
In \cite{av_2}, Wang et.~al proposed a convolution neural network based autoencoder for anomaly detection where the autoencoder detects the anomaly by comparing the reconstruction error with a pre-defined threshold. At the end of the paper, the authors provided a transfer learning scheme, that is, first train an autoencoder with a large airport's data, and then train on other airports' data by freezing the first layer of the original of the autoencoder. However, the authors did not further investigate the scheme and therefore no result is shown. 

Consequently, how to reduce total training time and obtain high-performance data-driven models when data is scarce for data-driven air traffic applications has not been investigated. In addition, the multi-agent property of the ATM system cannot be explicitly taken into account with the existing Transformer-based models that have been utilized for pre-training and fine-tuning in other domains. In response to these research gaps, we propose the Multi-Agent BERT (MA-BERT) which replaces the multi-head attention in BERT with the agent-aware attention \cite{agent}. We also propose to apply the MA-BERT on a pre-training and fine-tuning framework for data-driven air traffic applications. With the agent-aware attention, MA-BERT can focus on learning the air traffic controllers' decisions during pre-training, similar to how BERT learns human’s understanding of natural language. This leads to higher performance compared to not only BERT, but also recurrent neural networks such as sequence-to-sequence LSTM. By applying MA-BERT to the pre-training and fine-tuning framework, a large amount of total training time can be saved and the pre-trained MA-BERT can be quickly adapted to airports with small datasets or even newly constructed airports with no historical data at all. 

The main contributions of this paper are: (i) demonstration of how to reduce total training time and obtain high-performance data-driven models when data is scarce for data-driven air traffic applications; (ii) a novel multi-agent based transfer learning model, MA-BERT, that explicitly takes into account the multi-agent property of ATM system; and (iii) demonstration of the feasibility of the pre-training and fine-tuning framework for data-driven air traffic applications. 

The rest of this paper is organized as followed: Sec.~\ref{sec:ma_bert} presents the problem formulation and explains the proposed MA-BERT. Sec.~\ref{sec:pre} and \ref{sec:fine} explains the pre-training and fine-tuning scheme in details. Sec.~\ref{sec:result} describes the data being used to validate the MA-BERT, experiment details, and the testing results. The final conclusion and limitations are given in Sec.~\ref{sec:con}.

\section{Methodology} \label{sec:method}
\begin{figure}[htbp]
\centering
\includegraphics[width=1.0\linewidth]{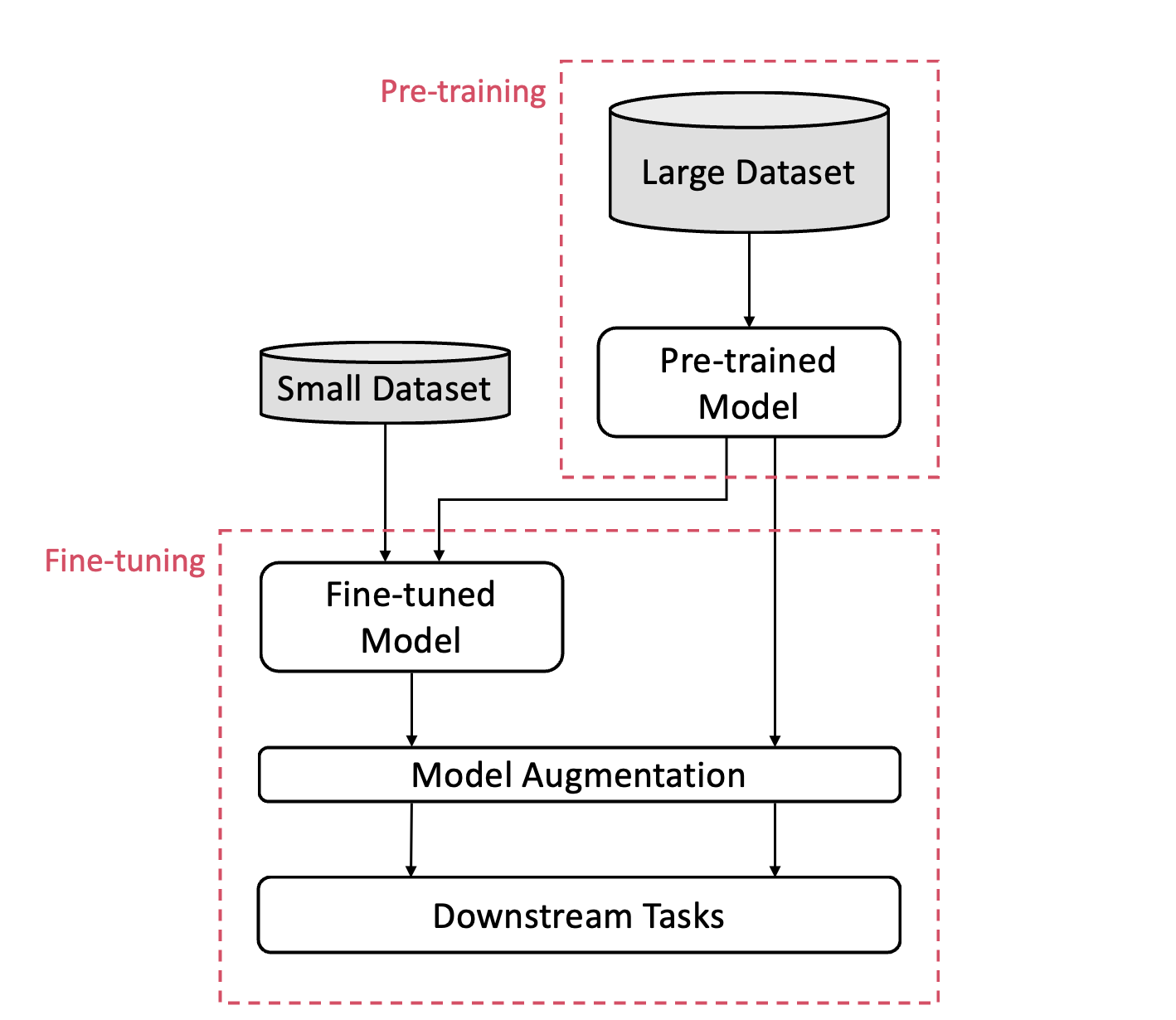}
\caption{General framework of pre-training and fine-tuning}
\label{fig:flowchart}
\end{figure}

The general framework of pre-training and fine-tuning is shown in Fig.~\ref{fig:flowchart}. Firstly, a pre-trained model is obtained by training on a large dataset such as the dataset from a major airport collected for an entire year. Then, for downstream tasks of an existing minor airport with small dataset, the pre-trained model can be further trained for just a few epochs to obtain a fine-tuned model. The fine-tuned model is then augmented by adding another component to adapt to different downstream tasks. For a newly constructed airport that does not have any dataset, the pre-trained model can be directly augmented and updated regularly for downstream tasks. The rest of this the section explains the structure of the proposed Multi-Agent Bidirectional Encoder Representations from Transformers (MA-BERT), and the pre-training and fine-tuning strategies in details. 

\subsection{MA-BERT} \label{sec:ma_bert}
\begin{figure}[htbp]
\centering
\includegraphics[width=0.9\linewidth]{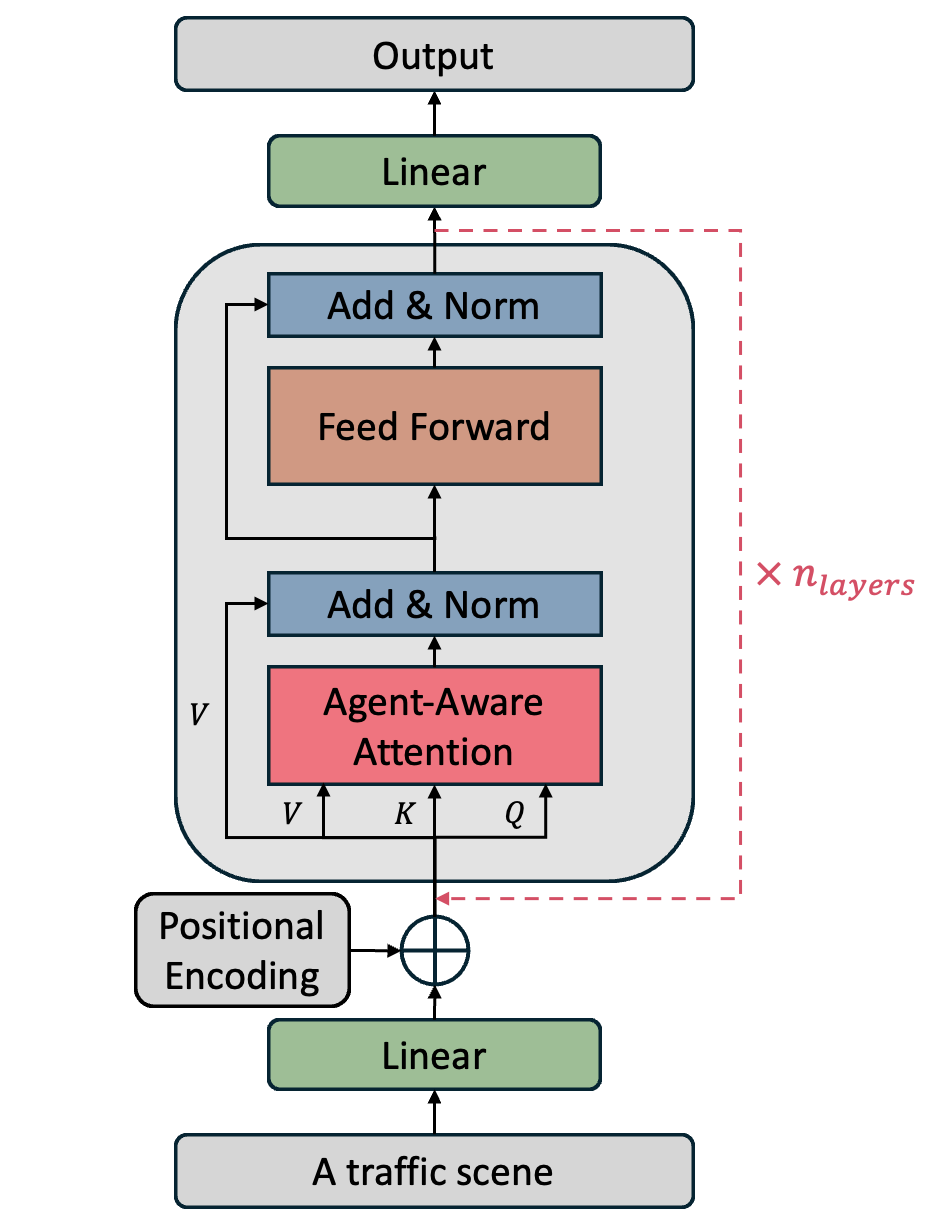}
\caption{Proposed MA-BERT structure}
\label{fig:ma_bert}
\end{figure}

This section presents and explains the proposed MA-BERT. As shown in Fig.~\ref{fig:ma_bert}, MA-BERT is an variant of BERT that replaces the multi-head attention with the agent-aware attention proposed in \cite{agent}. In general, BERT is the encoder part of the Transformer proposed in \cite{transformer} and is proposed by Devlin et. al in 2018 \cite{bert} to be utilized for pre-training and fine-tuning of natural language tasks. The design of BERT's structure enables it to be bidirectionally trained on unlabeled text and learns the deep representation of languages. Thus, BERT can be pre-trained on large datasets and then fine-tuned by simply adding an extra output layer to create the state-of-the-art performance in 11 natural language processing tasks. Since the air traffic surveillance data such as the Automatic Dependent Surveillance-Broadcast (ADS-B) data are time series data, similar to natural languages, models such as BERT can be adopted to air traffic applications with small adjustments. However, air traffic management system is a multi-agent system where Air Traffic Controllers (ATCs) make decisions based on all the aircraft, i.e., traffic situations. Therefore, the interactions between aircraft cannot be ignored if we want to achieve high performance with the pre-training and fine-tuning framework. To do so, we propose MA-BERT by replacing the multi-head attention with the agent-aware attention. The agent-aware attention enables the MA-BERT to generate outputs based on the traffic situation rather than a single aircraft to achieve higher performance than the standard BERT.

The structure of the MA-BERT for pre-training is shown in Fig.~\ref{fig:ma_bert}. Rather than a single trajectory, MA-BERT is trained on every traffic scene. A traffic scene that has $T$ time steps, $N$ number of agents (aircraft), and $F$ number of features for each agents is denoted as $S \in \mathbb{R}^{N\times T\times F}$ and can be expressed as:
\begin{equation}\label{eqn:S}
S = \left[X^1_{T}, X^2_{T}, \cdots, X^n_{T}, \cdots, X^N_{T}\right] 
\end{equation}
where $n \in [1,N]$ and $X^n_{T}$ represents the $n^{th}$ agent, which can be expressed as:
\begin{equation}\label{eqn:X_t}
\begin{aligned}
X_{T}^n = &[\chi_{1,1}^n, \cdots, \chi_{1,F}^n, \cdots, \\
& \chi_{t_n,1}^n, \cdots, \chi_{t_n,F}^n, 0,\cdots,0]
\end{aligned}
\end{equation}
Note that every agent in scene $S$ has the same total number of features $F$ but different length. For instance, $t_n$ represents the total time steps of the $n^{th}$ agent in the scene. When $t_n$ is smaller than $T$, it means agent $n$ has reached the cutting point before the scene ends. The cutting point can be borders of a sector or final approach fixes of an airport. To allow agents with variable lengths to be expressed in the same scene, and to enable batch learning of multiple scenes with different lengths, we use padding and masks. We pad the data with zeros in the agents with smaller length, and express where the padded data is in the masks in the attention mechanism to let the model ignore the padded data. 

\begin{figure}[htbp]
\centering
\includegraphics[width=0.7\linewidth]{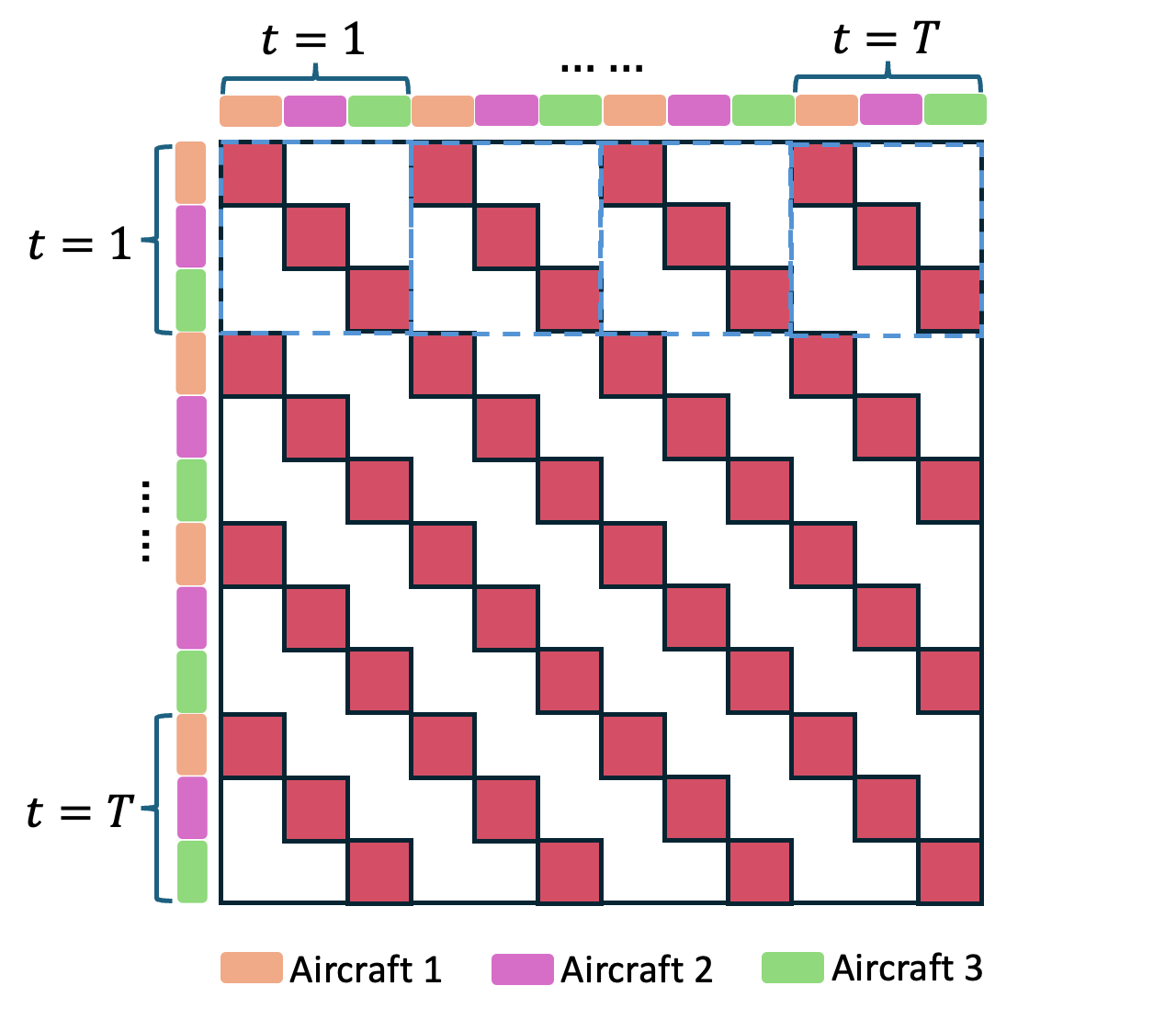}
\caption{An illustration of the mask in agent-aware attention}
\label{fig:M}
\end{figure}

\begin{figure}[htbp]
\centering
\includegraphics[width=1.0\linewidth]{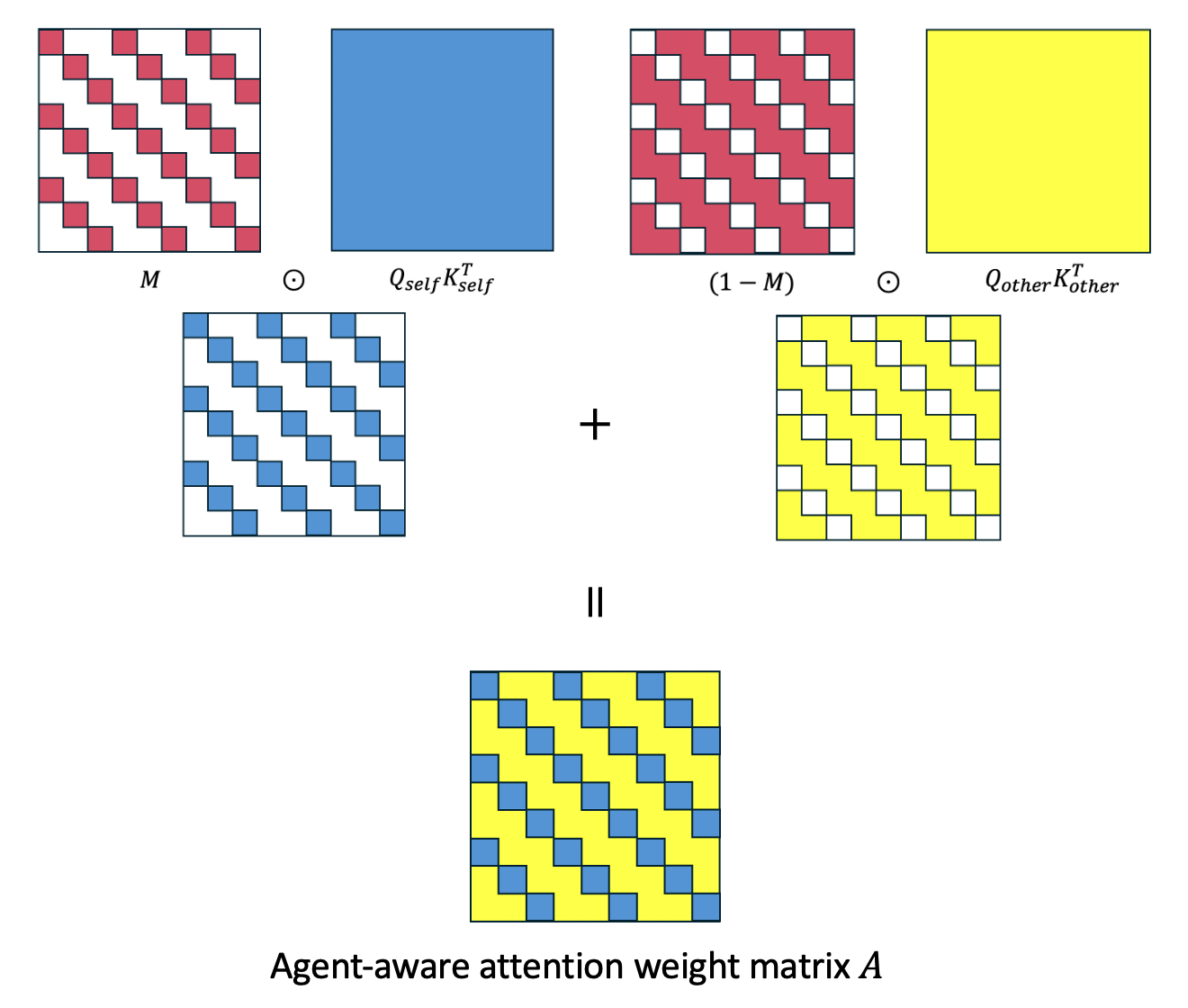}
\caption{An illustration of the agent-aware attention weight matrix}
\label{fig:A}
\end{figure}

Since MA-BERT, like all other transformer-based models, does not process time series data step by step like a Recurrent Neural Network (RNN), it requires a positional encoding to label each time step, which can be computed as:
\begin{equation} \label{eqn:pe}
PE(t,i) = 
    \begin{cases}
    \begin{aligned}
      &sin\left(\frac{t}{10000^{\frac{i}{d}}}\right), &&\text{if $i$ is even} \\
      &cos\left(\frac{t}{10000^{\frac{i-1}{d}}}\right), &&\text{if $i$ is odd}
    \end{aligned}
    \end{cases}
\end{equation}
where $t\in [0,T]$ is the time steps in the scene, $i\in [1,d]$, and $d$ is the dimension of the model \cite{transformer}. As shown in Fig.~\ref{fig:ma_bert}, MA-BERT first transforms $S$ into $\mathbb{R}^{N\times T\times d}$ with a linear layer, and then adds each agent's transformed states with the positional encoding based on their corresponding time steps in the scene. After the addition, MA-BERT takes the input and processes it with the agent-aware attention \cite{agent}. The agent-aware attention is computed as:
\begin{equation} \label{eqn:agent_attention}
\text{Agent-Aware Attention}(V,K,Q) = softmax\frac{A}{\sqrt{d}}V
\end{equation}
\begin{equation} \label{eqn:A}
A = M\odot(Q_{self}K^T_{self}) + (1-M)\odot(Q_{other}K^T_{other})
\end{equation}
\begin{align} 
Q_{self} = QW^Q_{self}, \;\;\;&K_{self} = KW^K_{self} \label{eqn:Qself}\\
Q_{other} = QW^Q_{other}, \;\;\;&K_{other} = KW^K_{other} \label{eqn:Qother}
\end{align}
where $V, K, Q \in \mathbb{R}^{N\cdot T \times d}$ are values, keys, and queries, respsectively, and they equal to the inputs. $\odot$ represents the element wise multiplication. $W_{self}^Q, W_{self}^K, W_{other}^Q, W_{other}^K \in \mathbb{R}^{d\times d}$ are trainable weights. In Eqn. \ref{eqn:A}, $M\in \mathbb{R}^{N\cdot T \times N\cdot T}$ is a binary mask that contains 1 for each agent themselves at each time step and 0 in all other places. An example of $M$ for a scene with 3 agents and 4 time steps is shown in Fig.~\ref{fig:M}. The red boxes in Fig.~\ref{fig:M} are 1 and all the white parts are 0. Blue boxes are examples of one time step. In this example, since there are 3 agents, each time step is represented as a $\text{3}\times\text{3}$ matrix. In addition, by using the agent-aware mask, the agent-aware attention weight matrix $A$ in Eqn. \ref{eqn:A} can be divided into 2 parts, an attention weight matrix for agents themselves and another attention weight matrix to other agents, as shown in Fig.~\ref{fig:A}. The agent-aware mask $M$ and the weight matrix $A$ help the agent-aware attention to distinguish each agent from themselves and force the MA-BERT to learn to pay attention to other agents. 

The output of the agent-aware attention is then added with the values $V$, followed by layer normalization, which has shown to significantly decrease training time and increase training stability \cite{layer_norm}. Then, the output from the layer normalization is passed into a feed-forward neural network that consists of several linear layers with the Rectified Linear Unit (ReLU) activation function \cite{relu}, which is also followed by an addition and layer normalization. The output of the second layer normalization is sent back to the beginning of the next layer, which has the same structure, and this process is iterated for $n_{layers}$ times before the last linear layer. As a result of that, the final output of MA-BERT is the input traffic scene's encoded representation $S' \in \mathbb{R}^{N\times T\times F}$ that can be expressed as:
\begin{equation} \label{eqn:s'}
S' = \left[Z^1_{T}, Z^2_{T}, \cdots, Z^n_{T}, \cdots, Z^N_{T}\right] 
\end{equation}
where $Z^n_{T}$ can be similarly expressed as:
\begin{equation}\label{eqn:Z_t}
\begin{aligned}
Z_{T}^n = &[\zeta_{1,1}^n, \cdots, \zeta_{1,F}^n, \cdots, & \\
&\zeta_{t_n,1}^n, \cdots, \zeta_{t_n,F}^n, \cdots, \zeta_{T,1}^n,\cdots,\zeta_{T,F}^n]
\end{aligned}
\end{equation}
Note that $Z_{T}^n$ has the same dimension as $X_{T}^n$. This is the key characteristics that enables MA-BERT to be connected to another output layer that transforms $Z_{T}^n$ into any dimension and fine-tuned to either forecast or classification tasks.

For comparison, we also implement the standard BERT and use it as a baseline model. BERT uses the multi-head attention that is proposed in for the Transformer model in \cite{transformer}. The multi-head attention is computed as:
\begin{equation} \label{eqn:mh_attention}
\begin{aligned}
\text{Multi-Head}&\text{ Attention}(V,K,Q) = \\
&\text{Concat}\left(H_1, H_2, \dots, H_h\right)W^{MH} \\
\text{where } H_j = &\text{Attention}(VW_j^V, KW_j^K, QW_j^Q)
\end{aligned}
\end{equation}
\begin{equation}
\text{Attention}(V,K,Q) = H = softmax\left(\frac{QK^T}{\sqrt{d}}\right)V \label{eqn:self_attention}
\end{equation}
where $h$ is the total number of heads, $j \in [1,h]$, $W^{MH} \in \mathbb{R}^{d \times d}$ and $W_j^V, W_j^K, W_j^Q \in \mathbb{R}^{d \times d_h}$ are trainable weights, and $d_h = \frac{d}{h}$. Unlike the proposed MA-BERT, the standard BERT is trained trajectory by trajectory. Thus, it takes a trajectory $X_T \in \mathbb{R}^{T\times F}$ that can be expressed as:
\begin{equation}
\begin{aligned}
X_T=&[\chi_{1,1}, \cdots, \chi_{1,F}, \cdots, \\
&\chi_{t,1}, \cdots, \chi_{t,F},\cdots, 0,\cdots,0 ]
\end{aligned}
\end{equation}
where $t \in [1,T]$. Similar to the MA-BERT, when $t$ is smaller than $T$, the rest of the trajectory is padded with zeros and a mask is utilized in the training to make sure the multi-head attention does not pay attention to the padded parts. The standard BERT outputs the deep representation of $X_t$, $Z_t \in \mathbb{R}^{T\times F}$ that can be expressed as:
\begin{equation}
\begin{aligned}
Z_T=&[\zeta_{1,1}, \cdots, \zeta_{1,F}, \cdots, \\
&\zeta_{t,1}, \cdots, \zeta_{t,F},\cdots, \zeta_{T,1}, \cdots, \zeta_{T,F} ]
\end{aligned}
  \end{equation}


\subsection{Pre-Training Scheme} \label{sec:pre}
\begin{figure}[htbp]
\centering
\includegraphics[width=1.0\linewidth]{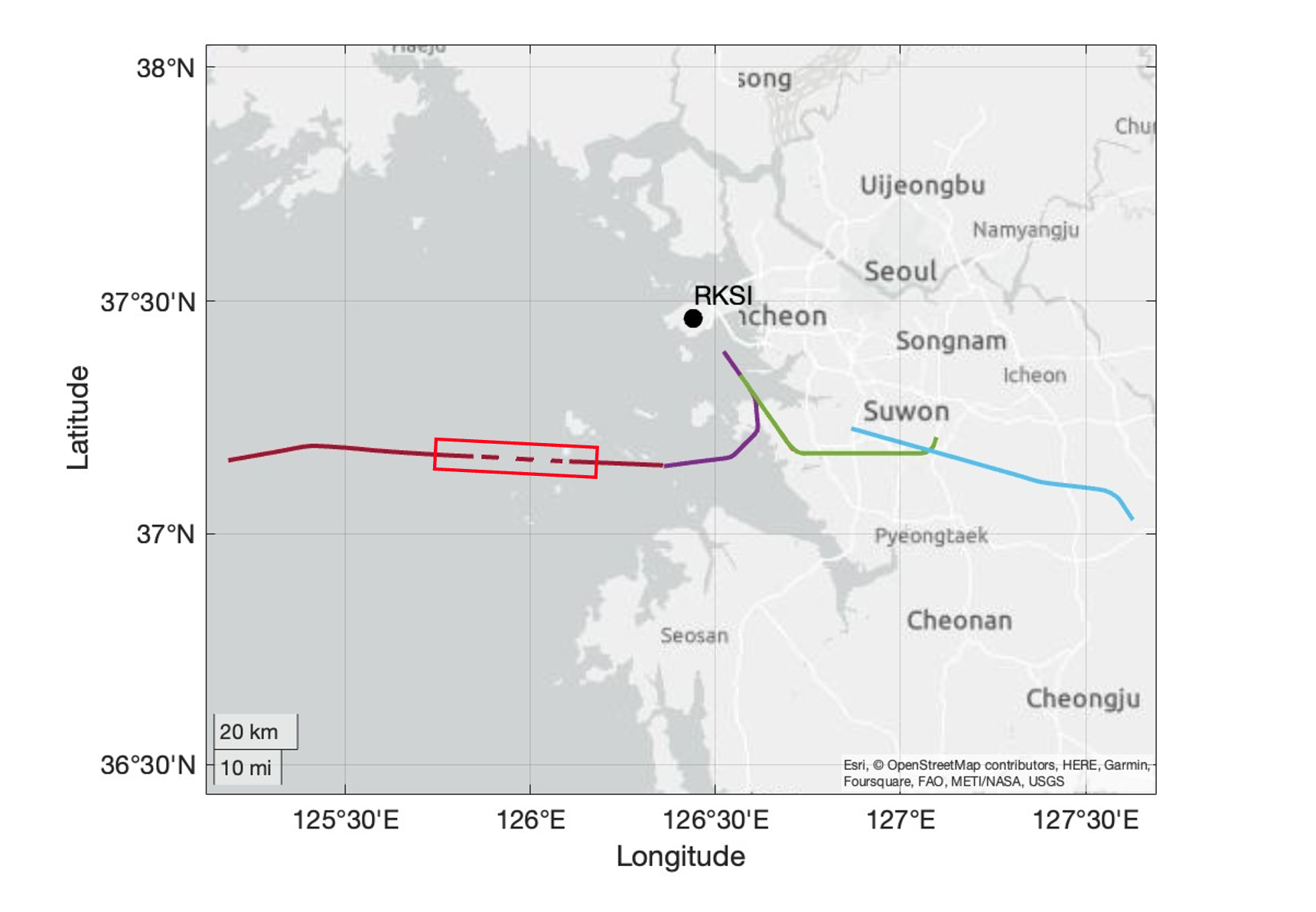}
\caption{An example of a masked scene for pre-training the MA-BERT: dashed parts are the masked parts}
\label{fig:mask_example}
\end{figure}

As shown in Figure \ref{fig:flowchart}, in order to fine-tune a model for downstream tasks, a pre-trained model is required. The pre-trained model is trained on a relatively large dataset such as the dataset from a major airport collected for an entire year. 
The pre-training process is a self-supervised training process and it requires the model to have an encoder structure. Therefore, the proposed MA-BERT and the baseline BERT are good candidates. As mentioned in Sec. \ref{sec:ma_bert}, the output of MA-BERT, $S'$, has the same shape of the input, $S$. 
The pre-training scheme for MA-BERT is masking parts of $S$, and making the model to recover the masked part through $S'$. The number of masks for each scene is 1 and the location of each mask is randomly selected so that the model can perform better after fine-tuned for downstream tasks \cite{multi_trans}. The length of the mask is 2 minutes, which correspond to the usual prediction length for trajectory prediction in terminal airspace \cite{mad_hs, HCchoi}. Note that any parts of a flight could be masked, which forces the MA-BERT to learn from the deep bidirectional representation during the pre-training \cite{bert}. In Fig.~\ref{fig:mask_example}, an example of a masked scene is shown. Each color represents a flight and the dashed part is the masked part. In this case, the dashed part of the red flight is masked. In pre-training, the MA-BERT's goal is to recover the dashed part based on the information of all agents. For pre-training the standard BERT, the same masking strategy is adopted except instead of masking a scene, a trajectory is masked. 

Let the masked parts of a given scene $S$ be $S_{masked}$ and the corresponding parts in $S'$ is $S'_{masked}$, the loss function for pre-training the MA-BERT is the Mean Squared Error (MSE) between $S_{masked}$ and $S'_{masked}$, which can be expressed as:
\begin{equation}\label{eqn:pre_loss}
\text{Pre-Training Loss} = MSE(S_{masked},S'_{masked})
\end{equation}
During pre-training, MA-BERT takes such masked scenes as inputs and it outputs an encoded representation of masked scenes where the masked parts are recovered. By doing so, MA-BERT learns the interactions between flights and ATCs' decisions. For the standard BERT, the loss function is the MSE between the masked parts in $X_T$ and the corresponding parts in $Z_T$. 

\subsection{Fine-Tuning Scheme} \label{sec:fine}
\begin{figure}[htbp]
\centering
\includegraphics[width=1.0\linewidth]{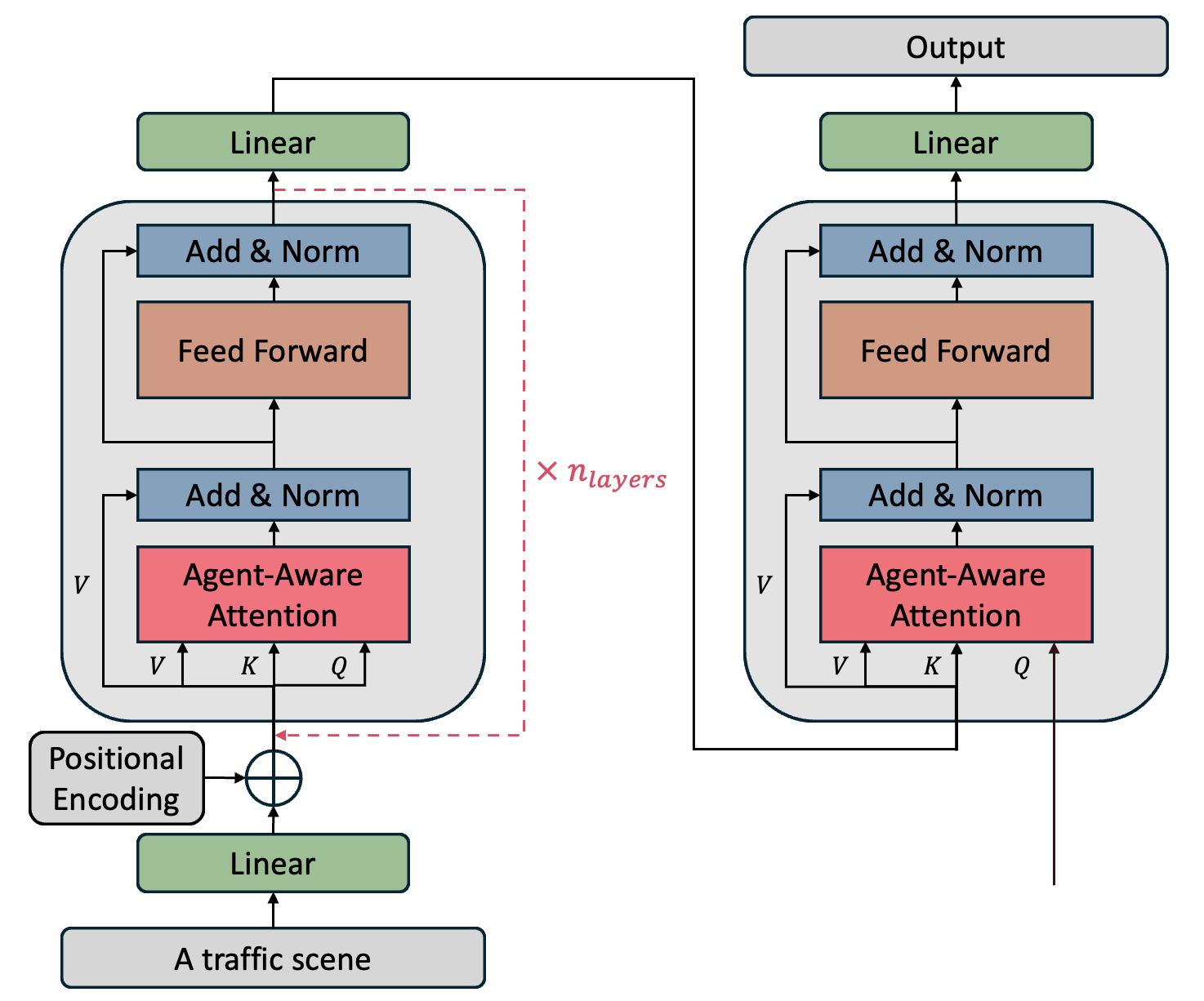}
\caption{Structure of augmented MA-BERT for fine-tuning}
\label{fig:fine_ma_bert}
\end{figure}

When the pre-training is finished, the model enters the fine-tuning stage. In this stage, the model can be either trained on another small dataset and then fine-tuned to downstream tasks or directly fine-tuned to downstream tasks. The first option can be done for small airports where some data has already been collected while the second option can be done for newly built or reconstructed airports. In addition, for the second option, we propose to implement an incremental learning strategy where the model is regularly updated so that it incrementally adapts the new dataset. For specific downstream tasks, we can augment the model to better suit the tasks. For instance, a decoder structure can be added for tasks such as ETA prediction. 
For tasks such as trajectory prediction, the model simply needs to change the masked parts to where the prediction is, and then take the masked parts from the output as the prediction. For the MA-BERT, the augmented model is shown in Fig.~\ref{fig:fine_ma_bert}. The augmented MA-BERT, compared to the MA-BERT in Fig.~\ref{fig:ma_bert}, has an additional decoder that takes the output from the encoder as the values and keys. Therefore, the values and keys in the decoder has the same dimension as the values and keys in the encoder. The queries in the decoder is a binary matrix in the dimension of $\mathbb{R}^{N\cdot O\times F}$ where $O$ is the dimension of the output for each agent and it is task specific. For instance, for Estimated Time of Arrival (ETA) prediction or binary classification tasks, $O$ equal to 1. The queries in the decoder equal to 1 for agents that need outputs and 0 for those that do not. For instance, for a scene that contains 5 flights but when the ETA prediction is made, only 2 of them are still in the airspace while the other 3 have landed, the query contains two ones and three zeros. This makes sure the model knows which agent needs an output. 

Note that a decoder is required for fine-tuning the MA-BERT because all the agents need to be permutation-invariant \cite{agent}. If only linear layers are added to transform the output from $\mathbb{R}^{N\times T\times F}$ to $\mathbb{R}^{O}$, the model can learn to generate outputs based on the order of the agents. For fine-tuning the standard BERT, another layer of BERT is used. The output is then sent to two linear layers to transform the output from $\mathbb{R}^{T\times F}$ to $\mathbb{R}^{O}$. For both the MA-BERT and the standard BERT, let the output from fine-tuning for each agent or trajectory be $\hat{Y} \in \mathbb{R}^{O}$ and the ground truth of the downstream task be $Y \in \mathbb{R}^{O}$, the loss function for fine-tuning is:
\begin{equation}\label{eqn:fine_loss}
\begin{aligned}
\text{Fine}&\text{-Tuning }\text{Loss} \\
&= 
\begin{cases}
	\begin{aligned}
		&MSE(\hat{Y}, Y), &&\text{Regression}\\
		&BCE(\hat{Y}, Y), &&\text{Binary Classification}\\
		&CCE(\hat{Y}, Y), &&\text{Multi-Class Classification}\\
	\end{aligned}
\end{cases}
\end{aligned}
\end{equation}
where $BCE$ is the binary cross entropy and $CCE$ is the categorical cross entropy. 

Since the pre-training has given the model to have a great amount of knowledge of how flights react to each other and how ATCs direct the flights in various traffic situations, the fine-tuning takes not only takes just a few epochs for each task, but also requires only a small amount of data to achieve even better performance than training the model for one specific task. 

\section{Results and Analysis} \label{sec:result}
In this section, we first present and describe the ADS-B data that is used to test our proposed MA-BERT and the transfer learning framework. For the downstream tasks, we select trajectory prediction and ETA prediction, which are the two fundamental air traffic applications. For comparison, we first include another model, the Sequence-to-Sequence Long Short-Term Memory (Seq2Seq LSTM), and then train all the models from scratch. The results are then compared with the results obtained from using the pre-trained models. It is shown that MA-BERT consistently performs better than the standard BERT and the Seq2Seq LSTM. It is also shown that using the transfer learning framework can not only improve the performance, but also save a large amount of total training time. In addition, the pre-trained MA-BERT is fine-tuned with different percentage of training data. The result shows the pre-trained MA-BERT requires only a small amount of data to achieve high performance. In the end, we simulate scenarios where MA-BERT is applied to new airports with no historical data available by using the incremental learning, i.e., updating the model regularly. 

\subsection{Data Description and Preparation} \label{sec:data}
The data used in this paper is the ADS-B data from three airports in South Korea, Incheon International Airport (RKSI), Gimpo International Airport (RKSS), and Gimhae International Airport (RKPK). RKSI is the largest airport in South Korea. Thus, it is used as the large dataset to train the pre-trained model. Since RKSS and RKSI share some fixes together, RKSS is used to test the performance of the transferred model from a major airport to a minor airport that operates similarly. RKPK is the smallest airport among them and it locates far from RKSI and RKSS. Thus, RKPK is used to test the performance of the transferred model in an airport that operates very differently. 

\begin{figure*}[htbp]
\centering
\includegraphics[width=1.0\linewidth]{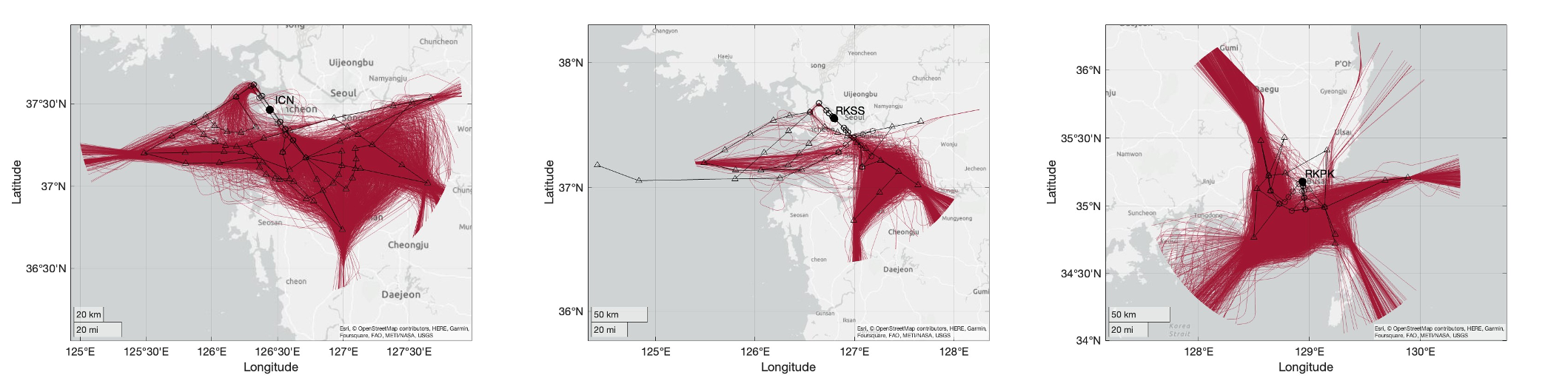}
\caption{Arrivals in January 2019 in RKSI (left), RKSS (middle), and RKPK (right)}
\label{fig:flights}
\end{figure*}

The ADS-B data recorded in 2019 is used. Among all features provided by the ADS-B data, the longitude, latitude, and altitude are used. The Aeronautical Information Publication (AIP) data is also used to find the FAFs. In this paper, only arrivals are considered because arrivals are much more complex than departures in the terminal airspace. 
All arrivals are firstly reconstructed using a regularized least-squares optimization \cite{preprocess}. 
\begin{equation}\label{eq:reconstruction}
    \underset{P}{minimize} \,\, \| CP - \tilde P \|^2_F + \lambda_2 \| D_2 P \|^2_F + \lambda_3 \| D_3 P \|^2_F 
\end{equation}
where given a flight with $t$ time steps, $P \in \mathbb{R}^{t \times 3}$ is the reconstructed trajectory, $C \in \mathbb{R}^{t \times t}$ is a diagonal matrix with its $i$'th element equal to one if the data was recorded at time $i$ and zero otherwise, and $\tilde{P} \in \mathbb{R}^{t \times 3}$ is the recorded trajectory filled with zero at times that were not recorded. $D_2$ and $D_3$ are the second and third-order difference matrices, representing the acceleration and jerk operator, respectively. $\lambda_2$ and $\lambda_3$ are positive tunable weighting parameters and are selected to smooth the reconstructed trajectory while keeping the reconstruction error low. The cost function in Eqn.~\ref{eq:reconstruction} simultaneously minimizes the reconstruction error and the change in velocity and acceleration (i.e., smoothness). The arrivals are then evenly sampled every 10 seconds and cut at 70 nautical miles away from the airports. Go-arounds and flights that conduct a holding command more than once are eliminated from the datasets. All arrivals in January 2019 after reconstruction and preprocessing in RKSI, RKSS, and RKPK are shown in Fig.~\ref{fig:flights}. The black lines with triangle markers and circle markers in Fig.~\ref{fig:flights} are the standard terminal arrival routes and instrument approach procedures, respectively, and the markers represent waypoints. The statistics of arrivals in all three airports is shown in Table~\ref{tab:stat}. We limit the maximum length of the traffic scenes as 10 minutes. Having a longer scene length is better for performance, but is limited to the memory size of the GPU when training. 
Note that for training the MA-BERT, scenes are used as the input instead of trajectories so for a fair comparison, all trajectories are the same trajectories in scenes. All datasets are spit in to 80\% for training, 10\% for validation, and 10\% for testing. 

\begin{table}[ht]
	\caption{Number of trajectories and scenes in each airport}
	\centering
	\begin{tabular}{c c c c}
		\toprule
		Data Types & RKSI & RKSS & RKPK \\
		\midrule
		Trajectories & 418,234 & 154,756 & 110,983 \\
		Scenes & 48,655 & 33,641 & 34,575 \\
		\bottomrule
	\end{tabular}
	\label{tab:stat}
\end{table}

\subsection{Downstream Tasks and Evaluation Metrics} \label{sec:task}
Two downstream tasks are selected in this paper: trajectory prediction and ETA prediction. The selected tasks are crucial applications in terminal airspace as they provide further information needed for many other applications that assist ATC's decision making. For trajectory prediction, the evaluation metrics are the Horizontal Error (HE) and the Vertical Error (VE), which are widely used metrics for trajectory prediction \cite{HCchoi,mad_hs}. HE is the distance between the flight's true and predicted horizontal position. Similarly, VE is the distance between the flight's true and predicted vertical position. For ETA prediction, the evaluation metrics are the Mean Absolute Error (MAE) and Root Mean Square Error (RMSE) between the true and predicted ETA. 

\subsection{Neural Network Setting} \label{sec:setting}
For training MA-BERT and BERT, several hyperparameter are required. For a fair comparison, they are set as the same for both models except the mini-batch size due to the large difference in sample size as shown in Table \ref{tab:stat}. Values of all hyperparameters are shown in Table \ref{tab:parameters}. The only hyperparameter for Seq2Seq LSTM is the dimension at each LSTM layer. The dimension first starts at 512 and then gradually decrease to encode the input. After that, the dimension gradually changes to 3 for trajectory prediction (longitude, latitude, and altitude) and 1 for ETA prediction. We keep the number of trainable parameters similar for each model. The total number of trainable parameter is 2,388,483 for MA-BERT and 2,380,291 for BERT for pre-training and trajectory prediction. After adding the decoder for fine-tuning for ETA prediction, the number of trainable parameter increases to  2,543,492 and 2,532,801 for MA-BERT and BERT. For Seq2Seq LSTM, the number of trainable parameter for trajectory prediction is 2,443,324 and 2,597,437 for ETA prediction. 
The difference in the number of trainable parameters is due to the difference in the agent-aware attention and the multi-head attention, but since the difference is less then 0.5\%, the effect is minor. The optimizer is the Adam optimizer \cite{adam} with learning rate set as $1\times 10^{-4}$ for pre-training and training new models, $2\times 10^{-5}$ and $5\times 10^{-5}$ for fine-tuning for trajectory prediction and ETA prediction, respectively. The learning rates are set through trial and errors and the best performing ones are selected. The models are built with Pytorch \cite{pytorch} in Python and ran on Google Colab with Nvidia A100 as GPU and Intel Xeon 2.2 GHz as CPU. 

\begin{table}[ht]
	\caption{Hyperparameters of MA-BERT and BERT}
	\centering
	\begin{tabular}{c c c c}
		\toprule
		Hyperparameters & Values \\
		\midrule
		Dimension of the model ($d$) & 512 \\
		Dimension of the feed-foward neural network & 2,048 \\
		Number of layers ($n_{layers}$) & 5 \\
		Number of heads ($h$) & 8 \\
		Dropout for pre-training & 0.1 \\ 
		Dropout for fine-tuning & 0 \\
		Mini-batch size for MA-BERT & 8 \\
		Mini-batch size for BERT & 16 \\
		\bottomrule
	\end{tabular}
	\label{tab:parameters}
\end{table}

\subsection{Model Comparison} \label{sec:compare}
To show the superiority of the proposed MA-BERT, we compare it with the standard BERT and Seq2Seq LSTM. As mentioned in Sec.~\ref{sec:setting}, the number of trainable parameters for all models are kept similar for a fair comparison. All models are trained from scratch for 100 epochs for trajectory prediction and ETA prediction. The results are shown in Table \ref{tab:all_performance} where the best results are highlighted in bold. Note that sec represents seconds, nm represents nautical miles, and ft represents feet. From Table \ref{tab:all_performance}, it can bee seen that BERT outperforms Seq2Seq LSTM in ETA prediction and has comparable performance in trajectory prediction, indicating the advantages of the multi-head attention. The proposed MA-BERT consistently performs the best. Compared to BERT, MA-BERT has a 15.77\% decrease in all metrics on average and 24.10\% decrease compared to Seq2Seq LSTM even though the sizes of all models are similar. This result shows the superiority of the proposed MA-BERT. 

\begin{table*}[ht]
\setlength\extrarowheight{3pt}
\caption{Performance comparison of MA-BERT, BERT, and LSTM}
\centering
\begin{tabular}{cccccc}
\toprule \vspace{0ex}
\multirow{2}{*}{Airports} & \multirow{2}{*}{Applications}          & \multirow{2}{*}{Metrics} & \multicolumn{2}{c}{Single-Agent} & Multi-Agent     \\ \cline{4-6}\vspace{0ex} 
                          &                                        &                          & Seq2Seq LSTM        & BERT         & MA-BERT         \\
\midrule
\multirow{4}{*}{RKSI}     & \multirow{2}{*}{ETA Prediction}        & MAE (sec)                &      76.29       & 71.36 & \textbf{66.33}  \\
                          &                                        & RMSE (sec)               &      127.6       &  121.6 & \textbf{101.3}\\
                          & \multirow{2}{*}{Trajectory Prediction} & HE (nm)  &      0.6001       & 0.6757 & \textbf{0.5485} \\
                          &                                        & VE (ft)                  & 261.0 & 259.2 & \textbf{227.5} \\
\midrule
\multirow{4}{*}{RKSS}     & \multirow{2}{*}{ETA Prediction}        & MAE (sec)                & 64.00             & 51.40        & \textbf{46.36}  \\
                          &                                        & RMSE (sec)               & 107.3            & 97.42        & \textbf{85.44}  \\
                          & \multirow{2}{*}{Trajectory Prediction} & HE (nm)                  & 0.6214            & 0.6735       & \textbf{0.4752} \\
                          &                                        & VE (ft)                  & 235.0             & 233.1        & \textbf{179.2}           \\
\midrule
\multirow{4}{*}{RKPK}     & \multirow{2}{*}{ETA Prediction}        & MAE (sec)                & 122.2             & 103.4        & \textbf{93.03}           \\
                          &                                        & RMSE (sec)               & 177.4             & 167.4        & \textbf{154.1}           \\
                          & \multirow{2}{*}{Trajectory Prediction} & HE (nm)                  & 0.7473            & 0.8201       & \textbf{0.6570}          \\
                          &                                        & VE (ft)                  & 276.0             & 306.3        & \textbf{239.3}     \\
\bottomrule     
\end{tabular}
\label{tab:all_performance}
\end{table*}

\begin{figure}[htbp]
\centering
\includegraphics[width=0.7\linewidth]{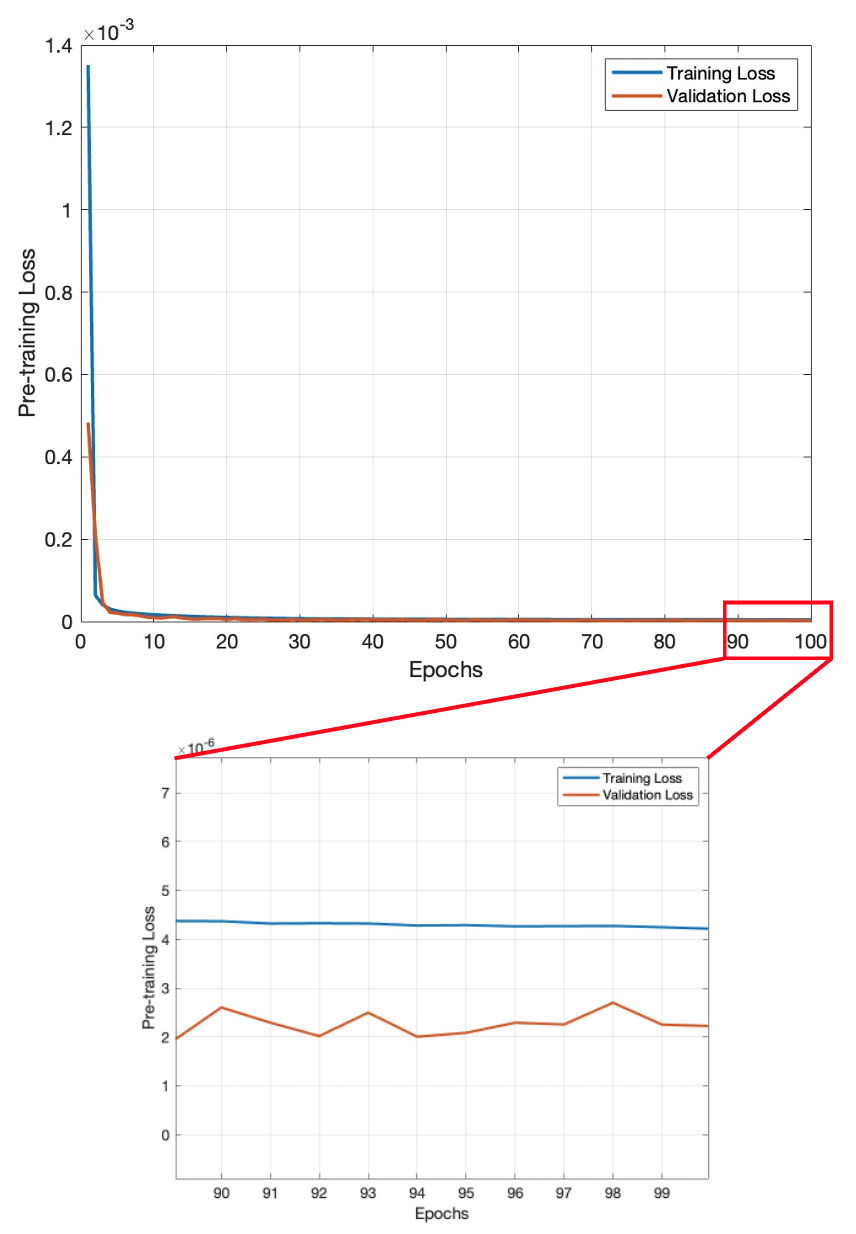}
\caption{Pre-training loss using the RKSI dataset}
\label{fig:pre_train_loss}
\end{figure}

\subsection{Fine-Tuning Results} \label{sec:fine_tuning}
The proposed MA-BERT can be used as the model for the pre-training and fine-tuning framework as shown in Fig.~\ref{fig:flowchart}. The data recorded in RKSI in 2019 is used as the big dataset for pre-training the MA-BERT. The pre-training loss is shown in Fig.~\ref{fig:pre_train_loss}. We use 100 epochs for pre-training. Each epoch takes about 671.0 seconds for MA-BERT. It can be seen that both the training and validation loss have converged and there is no overfitting, indicating a proper pre-training result. 


\begin{table*}[ht]
\setlength\extrarowheight{3pt}
\caption{Performance comparison of MA-BERT}
\centering
\begin{tabular}{cccccc}
\toprule \vspace{0ex}
\multirow{2}{*}{Airports} & \multirow{2}{*}{Applications}          & \multirow{2}{*}{Metrics} & New MA-BERT & Fine-tuned MA-BERT     \\ 
     &     &    & trained from scratch & for 10 Epochs \\
\midrule
\multirow{4}{*}{RKSI}     & \multirow{2}{*}{ETA Prediction}        & MAE (sec)                &      66.33       & \textbf{65.32}  \\
                          &                                        & RMSE (sec)               &     101.3        & \textbf{90.09}  \\
                          & \multirow{2}{*}{Trajectory Prediction} & HE (nm)                  &     0.5485       &  \textbf{0.5322} \\
                          &                                        & VE (ft)                  & 227.5 & \textbf{227.0} \\
\midrule

\multirow{4}{*}{RKSS}     & \multirow{2}{*}{ETA Prediction}        & MAE (sec)                & 46.36             & \textbf{44.13}  \\
                          &                                        & RMSE (sec)               & 85.44            & \textbf{82.64}  \\
                          & \multirow{2}{*}{Trajectory Prediction} & HE (nm)                  & 0.4752            & \textbf{0.4704} \\
                          &                                        & VE (ft)                  & 179.20 &\textbf{179.18} \\
\midrule
\multirow{4}{*}{RKPK}     & \multirow{2}{*}{ETA Prediction}        & MAE (sec)                & 93.03 & \textbf{86.63}           \\
                          &                                        & RMSE (sec)               & 154.1             & \textbf{139.6}       \\
                          & \multirow{2}{*}{Trajectory Prediction} & HE (nm)                  & 0.6570 & \textbf{0.6544}          \\
                          &                                        & VE (ft)                  & 239.3             & \textbf{239.0}     \\
\bottomrule     
\end{tabular}
\label{tab:tl_performance}
\end{table*}

To test the performance of the fine-tuned MA-BERT, we first pre-train a MA-BERT with the large RKSI dataset for 100 epochs and then fine-tune with RKSS and RKPK datasets for 10 epochs, as illustrated in Fig.~\ref{fig:flowchart}. 
For downstream tasks, a decoder is added for the ETA prediction task and the entire model is trained for 10 more epochs for each task. Since during the pre-training, the model is already converged, as shown in Fig.~\ref{fig:pre_train_loss}, only a few epochs is enough for fine-tuning. For example, 3 epochs are used for fine-tuning the BERT in the original paper \cite{bert}. We choose 10 in this paper as an illustration. More or less epochs can be chosen if needed. 
Table \ref{tab:tl_performance} shows the performance results where the best results are also highlighted in bold. It can be seen that the fine-tuned MA-BERT outperforms the MA-BERT that is trained from scratch with each airport's data. Especially for ETA prediction in RKPK, the fine-tuned MA-BERT has a 8.144\% decrease in MAE and RMSE on average. Note that the pre-trained MA-BERT is fine-tuned with only 20 epochs, which saves a large amount of total training time. Since for MA-BERT, each epoch takes about 671.0 seconds for RKSI's data, 235.4 seconds for RKSS's data and 182.5 seconds for RKPK's data, total training time for training trajectory prediction and ETA prediction models in all airports takes about 60.50 hours. However, the total training time for getting the same models by using the transfer learning framework only takes about 25.85 hours. The statistics are summarized in Table \ref{tab:time_comparison}. Note that the total training time for new MA-BERT and fine-tuned MA-BERT is the time for obtaining each model for both ETA prediction and trajectory prediction. In total, 34.65 hours of training time are saved. Note that with the increase of more tasks and airports, training time can be much further reduced. Thus, the test results show our proposed MA-BERT outperforms the standard BERT and Seq2Seq LSTM, and by incorporating MA-BERT into the transfer learning framework, not only the performance is improved, but also a large amount of training time is saved. 

\begin{table*}[ht]
\setlength\extrarowheight{3pt}
\caption{Total training time comparison between new MA-BERT and fine-tuned MA-BERT}
\centering
\begin{tabular}{cccc}
\toprule \vspace{0ex}
\multirow{2}{*}{Airports} & \multirow{1}{*}{Training time } & \multirow{1}{*}{Total training time} & \multirow{1}{*}{Total training time} \\ & of 1 epoch (sec) & for new MA-BERT (hour) & for fine-tuned MA-BERT (hour) \\
\midrule
\multirow{2}{*}{RKSI}  & \multirow{2}{*}{671.0} & \multirow{2}{*}{37.28} & \multirow{2}{*}{22.37} \\  & & & \\
\midrule
\multirow{2}{*}{RKSS}     &  \multirow{2}{*}{235.4} & \multirow{2}{*}{13.08} & \multirow{2}{*}{1.962} \\ & & & \\
\midrule
\multirow{2}{*}{RKPK}     &  \multirow{2}{*}{182.5} & \multirow{2}{*}{10.14} & \multirow{2}{*}{1.521} \\ & & & \\
\bottomrule     
\end{tabular}
\label{tab:time_comparison}
\end{table*}

Another important application of the pre-trained model is its ability to quickly adapt to another dataset. To show this, we apply a MA-BERT that has been pre-trained on RKSI's data and we fine-tune it to only certain parts of the data from RKSS and RKPK. More specifically, we fine-tune the pre-trained MA-BERT on 20\%, 40\%, 60\%, and 80\% of the training data from RKSS and RKPK for 10 epochs. The testing data is kept the same for a fair comparison with the result in Table~\ref{tab:all_performance} and \ref{tab:tl_performance}. It can be seen that fine-tuning the pre-trained MAT-BERT with only 20\% of the training data can already deliver comparable performance compared to Seq2Seq LSTM and the standard BERT trained with all of the training data for 100 epochs. 

\begin{table*}[ht]
\setlength\extrarowheight{3pt}
\caption{Performance comparison of MA-BERT fine-tuned with different percentage of data}
\centering
\begin{tabular}{ccccccc}
\toprule \vspace{0ex}
\multirow{2}{*}{Airports} & \multirow{2}{*}{Applications}          & \multirow{2}{*}{Metrics} & \multicolumn{4}{c}{Percentage of Available Training Data}   \\ \cline{4-7}\vspace{0ex} 
                          &                                        &                          & 20\%       & 40\%         & 60\% & 80\%         \\
\midrule
\multirow{4}{*}{RKSS}     & \multirow{2}{*}{ETA Prediction}        & MAE (sec)                &      61.95       & 52.76 & 51.22 & \textbf{49.01}  \\
                          &                                        & RMSE (sec)               &      96.81       & 98.72 & 95.38 & \textbf{89.53} \\
                          & \multirow{2}{*}{Trajectory Prediction} & HE (nm)  &        0.6064     & 0.5453 & 0.5076 & \textbf{0.4951} \\
                          &                                        & VE (ft)                  & 190.7 & 186.4 & \textbf{183.5} & 190.5 \\
\midrule
\multirow{4}{*}{RKPK}     & \multirow{2}{*}{ETA Prediction}        & MAE (sec)                & 155.7             & 111.2        & 107.0 & \textbf{104.5}          \\
                          &                                        & RMSE (sec)               & 198.6             & 162.6        & 159.4 & \textbf{157.6}           \\
                          & \multirow{2}{*}{Trajectory Prediction} & HE (nm)                  & 0.8452            & 0.7692       & 0.7329 & \textbf{0.7274}         \\
                          &                                        & VE (ft)                  & 255.5             & \textbf{246.2}        & 275.6  & 256.0   \\
\bottomrule     
\end{tabular}
\label{tab:percent}
\end{table*}

\begin{table*}[ht]
\setlength\extrarowheight{3pt}
\caption{Performance comparison of MA-BERT under different update frequencies}
\centering
\begin{tabular}{cccccc}
\toprule \vspace{0ex}
\multirow{2}{*}{Airports} & \multirow{2}{*}{Applications}          & \multirow{2}{*}{Metrics} & \multicolumn{3}{c}{Incremental Learning Update Frequency}   \\ \cline{4-6}\vspace{0ex} 
                          &                                        &                          & Every Day       & Every Week         & Every Month         \\
\midrule
\multirow{4}{*}{RKSS}     & \multirow{2}{*}{ETA Prediction}        & MAE (sec)                &      60.04       & \textbf{51.00} & 51.96  \\
                          &                                        & RMSE (sec)               &      108.8       & \textbf{95.25} & 96.08 \\
                          & \multirow{2}{*}{Trajectory Prediction} & HE (nm)  &      0.6673       & 0.5573 & \textbf{0.5441} \\
                          &                                        & VE (ft)                  & 226.7 & 198.3 & \textbf{198.0} \\
\midrule
\multirow{4}{*}{RKPK}     & \multirow{2}{*}{ETA Prediction}        & MAE (sec)                & 113.0             & 104.9        & \textbf{102.4}           \\
                          &                                        & RMSE (sec)               & 184.5             & 175.1        & \textbf{169.4}           \\
                          & \multirow{2}{*}{Trajectory Prediction} & HE (nm)                  & 1.000            & 0.9202       & \textbf{0.8761}          \\
                          &                                        & VE (ft)                  & 282.7             & 267.8        & \textbf{258.2}     \\
\bottomrule     
\end{tabular}
\label{tab:il_performance}
\end{table*}

\subsection{Incremental Learning Results} \label{sec:incremental_result}

The pre-trained model's ability to quickly adapt to another dataset means it can be applied to newly build airports where no data is available. Therefore, we adopt the idea of incremental learning and simulate such scenarios by using the pre-trained MA-BERT and updating it regularly with RKSS and RKPK's data. The performance summary is shown in Table~\ref{tab:il_performance}. Note that RKSI is not shown since RKSI's data is used for pre-training MA-BERT and it cannot be used for incremental learning anymore. 
It can be seen that for both ETA prediction and trajectory prediction, updating the model every month delivers the lowest error except for the ETA prediction in RKSS. Updating every day, however, makes the performance much lower. The reason to this could be that one day's traffic volume is too low to provide enough data to train the model. RKSS and RKPK have only 193 and 144 flights on average every day. Thus, the model is prone to overfitting when it is updated every day. Testing with variable number of epochs and learning rate could potentially alleviate the overfitting. Although updating every week has the lowest error on ETA prediction in RKSS, the difference compared to updating every month is only about 1-2\%, which could be the result of hyperparameter tuning. The incremental learning result shows the pre-trained MA-BERT, when applied to airports with no historical data, can still achieve higher performance than the  Seq2Seq LSTM and standard BERT, which further shows the effectiveness of MA-BERT and the pre-training and fine-tuning framework.

\section{Conclusion} \label{sec:con}
In this paper, Multi-Agent Bidirectional Encoder Representations from Transformers (MA-BERT) has been proposed and it has been applied to the pre-training and fine-tuning framework for data-driven air traffic applications. The MA-BERT replaces the multi-head attention in the standard BERT with the agent-aware attention, giving MA-BERT the ability to learn Air Traffic Controllers' (ATCs) decisions. The pre-training and fine-tuning framework provides a scheme for transferring a pre-trained model that is trained on a large dataset to smaller airports with small datasets or newly constructed airports with no historical data at all. After the transfer, the model can be fine-tuned to downstream tasks. 
The test data being used in this paper include the Automatic Dependent Surveillance-Broadcast (ADS-B) of three airports in South Korea, Incheon International Airport (RKSI), Gimpo International Airport (RKSS), and Gimhae International Airport (RKPK). Through experiments with the ADS-B data, the results have shown (i) MA-BERT's ability to achieve higher performance than the standard BERT and Sequence-to-Sequence Long Short-Term Memory (Seq2Seq LSTM) in trajectory prediction and Estimated Time of Arrival (ETA) prediction, (ii) the reduction of total training time through the use of the pre-training and fine-tuning framework, and (iii) the ability of the pre-trained MA-BERT to maintain its performance when it is transferred to new airports.

Note that there exist two limitations of this paper. First, only three airports in South Korea are tested. Thus, it remains unknown that how well will MA-BERT perform in airports in other countries. Second, domain information such as the arrival procedures are not used. It is known that ATCs' actions are highly connected to the arrival procedures. Thus, if such information can be incorporated into MA-BERT or the pre-training and fine-tuning framework, the performance may be able to be improved even more. The two limitations are considered as our future works and will be studied to further refine this work. 

\section*{Acknowledgments}
This work is supported by the Korea Agency for Infrastructure Technology Advancement (KAIA) grant funded by the Ministry of Land, Infrastructure and Transport (Grant RS-2020-KA158275).



\bibliography{references}
\bibliographystyle{IEEEtran}

%
 

\begin{IEEEbiography}[{\includegraphics[width=1in,height=1.25in,clip,keepaspectratio]{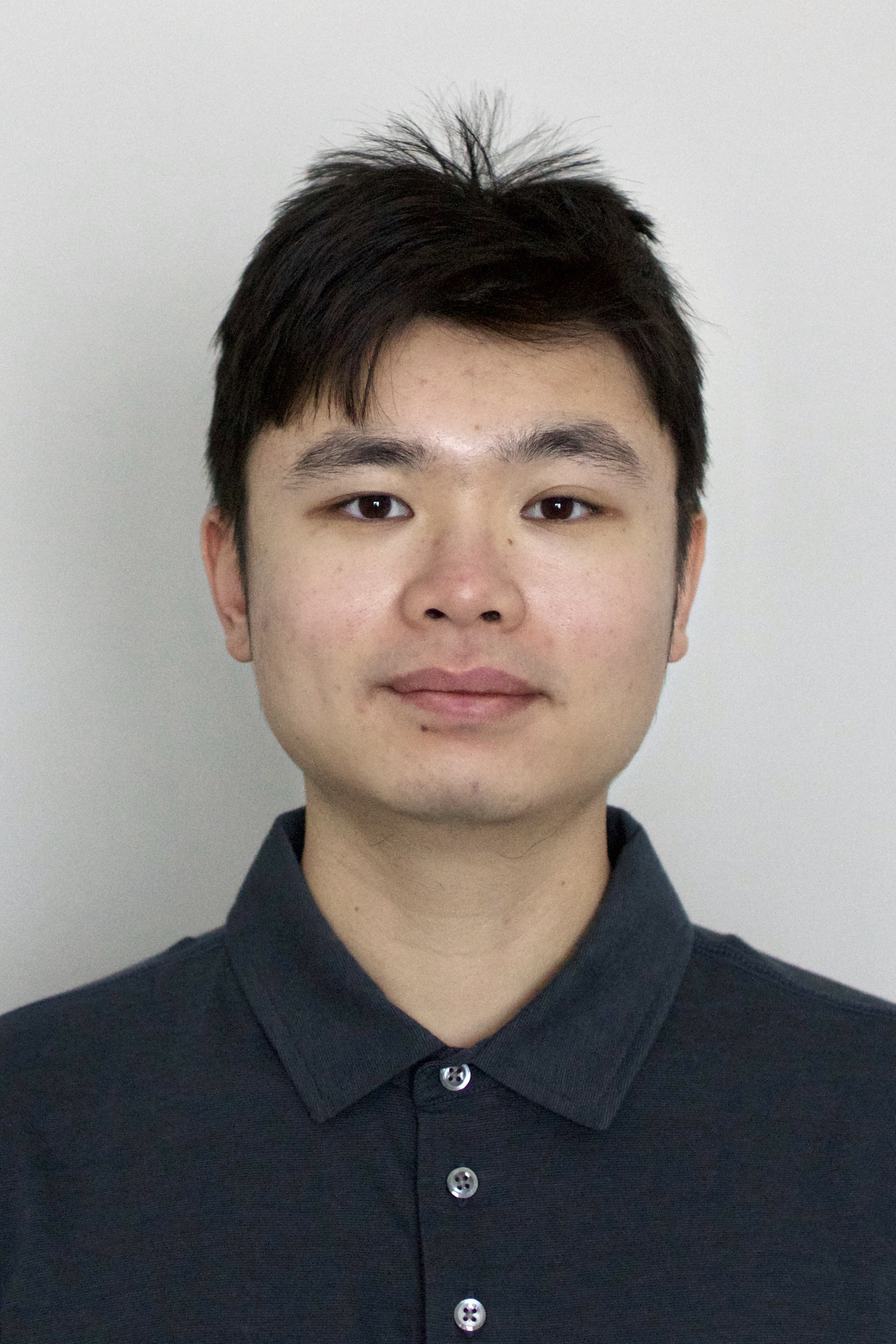}}]{Chuhao Deng}
received the B.S. and the M.S. degree from Purdue university. He is currently pursuing the Ph.D. degree in aeronautics and as- tronautics from Purdue University, West Lafayette, IN. 

His research interests include air traffic man- agement, human-machine interaction, and machine learning.
\end{IEEEbiography}

\begin{IEEEbiography}[{\includegraphics[width=1in,height=1.25in,clip,keepaspectratio]{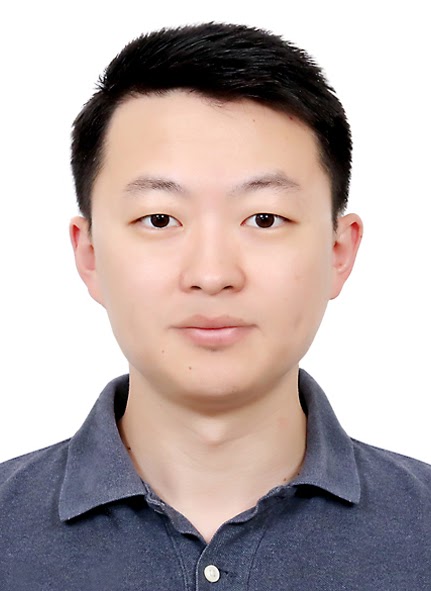}}]{Hong-Cheol Choi}
received the B.S. degree in applied physics from Korea Military Academy, Seoul, Korea, in 2012 and the M.S. degree in mechanical engineering from Seoul National Uni- versity, Seoul, Korea, in 2016. He is currently pursuing the Ph.D. degree in aeronautics and as- tronautics from Purdue University, West Lafayette, IN.

His research interests include machine learning, data mining, and air traffic management.
\end{IEEEbiography}

\begin{IEEEbiography}[{\includegraphics[width=1in,height=1.25in,clip,keepaspectratio]{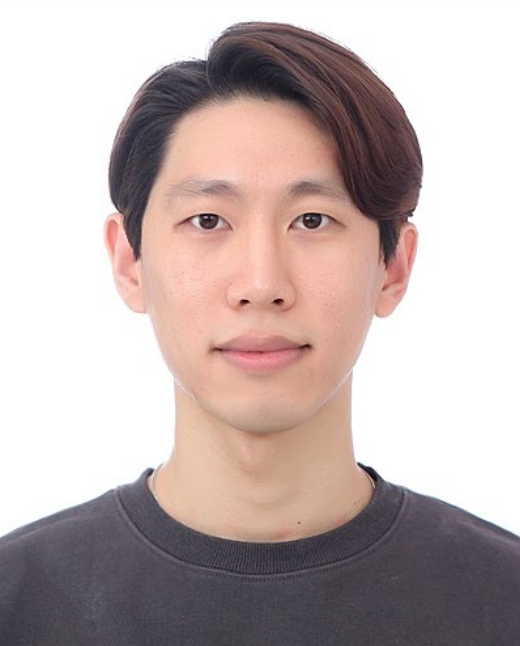}}]{Hyunsang Park}
received the B.S. and M.S. degree in mechanical and aerospace engineering from Seoul National University, Seoul, South Korea, in 2018 and 2020, respectively. He is currently working toward the Ph.D degree in aeronautics and astronautics with Purdue University, West Lafayette, IN, USA. 

His research interests include data-driven control, learning and control, air traffic management, and reachability analysis. 
\end{IEEEbiography}
\begin{IEEEbiography}[{\includegraphics[width=1in,height=1.25in,clip,keepaspectratio]{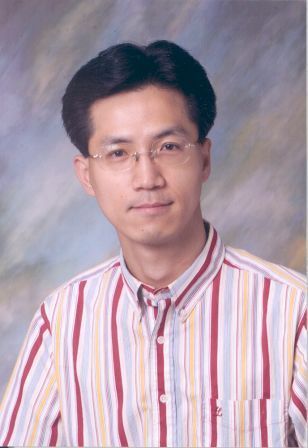}}]{Inseok Hwang}
received the B.S. degree from Seoul National University, Seoul, Korea, and the M.S. degree from Korea Advanced Institute of Sci- ence and Technology (KAIST), Daejeon, Korea, both in aerospace engineering, and the Ph.D. de- gree in aeronautics and astronautics from Stanford University, Stanford, CA, in 2004.

He is currently an Assistant Professor with the School of Aeronautics and Astronautics, Purdue University, West Lafayette, IN. He also holds an affiliate appointment with the system-of-systems signature area at Purdue University. His research interests include control and information inference of complex networked embedded systems such as transportation systems, networked robotics, communication, and sensor networks, and biological systems, and their application to multiple-vehicle systems, especially to air traffic surveillance and control. For his research, he leads the Flight Dynamics and Control/Hybrid Systems Laboratory, Purdue University.

Dr. Hwang was a recipient of the NSF CAREER Award on stochastic hybrid systems and its application to mobile networked embedded systems in 2008. He is currently an associate fellow of the AIAA, and a member of the IEEE Control Systems Society and Aerospace and Electronics Systems Society.
\end{IEEEbiography}



\vfill

\end{document}